\documentclass[letterpaper, 10 pt, conference]{ieeeconf}  
\IEEEoverridecommandlockouts 
\usepackage{graphics} 
\usepackage{epsfig} 
\usepackage{amsmath} 
\usepackage{amssymb}  
\usepackage{bm}
\usepackage{multirow}
\usepackage{graphicx}
\usepackage{float}
\usepackage{subfig}
\usepackage{url}
\usepackage{hyperref}
\hypersetup{hypertex=true,
hidelinks=true,
urlcolor=blue,
colorlinks=true,
linkcolor=blue,
anchorcolor=blue,
citecolor=blue}
\usepackage{flushend}
\usepackage{booktabs} 
\usepackage{bbding}
\usepackage{xcolor} 

\usepackage{soul} 
\usepackage[table]{xcolor} 

\usepackage{wrapfig}
\usepackage{tabularx}
\usepackage{algorithm}     
\usepackage{algpseudocode} 

\usepackage{tcolorbox}

\usepackage{tikz}     

\title{ \LARGE \bf
Assistron: Bayesian Shared Autonomy with Off-the-shelf Vision-Language-Action Models
}

\author{Pinhao Song$^{\dagger,1,3}$, Ze Fu$^{\dagger,1,3}$, Yutong Hu$^{1,3}$, Renaud Detry$^{1,2,3}$
\thanks{Supported by Interne Fondsen KU Leuven/Internal Funds KU Leuven (C14/22/134).}
\thanks{$^{1}$KU Leuven, Dept. Mechanical Engineering, Research unit \emph{Robotics, Automation and Mechatronics}, B-3000 Leuven, Belgium.
        {\tt\small Email: firstname.lastname@kuleuven.be}}
\thanks{$^{2}$KU Leuven, Dept. Electrical Engineering, Research unit \emph{Processing Speech and Images}, B-3000 Leuven, Belgium.
}
\thanks{$^{3}$Flanders Make@KU Leuven.}
\thanks{$^{\dagger}$ Equal contribution.}
}

\begin{document}

\maketitle

\begin{abstract}
We propose \emph{Assistron}, a shared autonomy model that leverages Vision-Language-Action (VLA) models to assist the user in daily activities. Our approach is grounded in two core principles: (1)~minimizing human cognitive and physical effort by leveraging VLA-driven autonomy for macro-movements, and (2)~prioritizing human intervention specifically at critical failure points. Driven by the user's verbal language commands, Assistron utilizes the VLA to autonomously execute macro-reaching trajectories, saving users' effort. In contact-rich interactions where VLAs tend to fail, Assistron employs a phase-aware interaction detection mechanism and solicits the user to intervene, in turn adjusting the VLA's action generation via flow matching guidance. Critically, our formulation eliminates the need for VLA fine-tuning, protecting its broad behavioral priors from catastrophic forgetting and ensuring the model does not become a narrow specialist.  We validate our approach on a comprehensive multi-task scene recovery benchmark encompassing diverse daily manipulation skills. Empirical results demonstrate that Assistron significantly improves task success rates over pure autonomous baselines while significantly reducing human cognitive and physical workload compared to traditional teleoperation, offering a scalable, smooth, and effortless paradigm for assistive manipulation. The code is available in \url{https://github.com/mousecpn/Assistron.git}.
\end{abstract}

\section{Introduction}
Assistive robotics aims to support individuals with motor impairments in activities of daily living. However, a long-standing challenge in assistive technology is the tension between system versatility and engineering complexity. Traditionally, assistive controllers are engineered for narrow, predefined tasks such as inserting~\cite{chang2021shared}, grasping~\cite{rt}, or pouring water~\cite{sct}. In contrast, everyday human environments demand general-purpose intelligence capable of executing an unpredictable variety of actions---from picking up a pill to sliding a drawer.

Vision-language-action (VLA) models have recently demonstrated remarkable open-world generalization across various manipulation tasks by leveraging internet-scale semantic priors and large-scale robot data~\cite{intelligence2025pi_}. These capabilities make VLA policies particularly attractive for assistive robotics, where robots must flexibly interpret diverse user intents and operate in unstructured environments. However, deploying VLA policies in assistive settings exposes a fundamental tension among precision, cost, and generality. Retraining a VLA to produce an assistive model demands an overwhelming amount of data \cite{ma2026aura, tang2026towards}. Instead of re-training a VLA, fine-tuning one requires less data, but it risks collapsing the broad behavioral priors of foundation policies into narrow, task-specific controllers \cite{cui2025end, liu2026adaptor}.  Ideally, an assistive model would use a VLA directly, without re-training or fine-tuning. However,  their straightforward use in assistive systems would unavoidably yield unsatisfactory performance, because of their relatively poor zero-shot reliability: Empirical evaluations, such as those in the RoboArena benchmark~\cite{atreya2025roboarena}, reveal that even state-of-the-art VLAs struggle to achieve success rates above 50\% (46.95\% and 35.25\% for $\pi_{0.5}$ and $\pi_{0}$, respectively\footnote{The statistics were derived from 667 demonstrations of $\pi_{0.5}$ and 1,125 demonstrations of $\pi_{0}$, all collected up to May 13th.}). 

Our work is built on the observation that the shortcomings in real-world manipulation are often not semantic failures, but localized execution failures concentrated around contact-rich interaction phases such as grasping, insertion, and object release. In many cases, the VLA correctly identifies the intended task and produces coherent long-horizon behavior, yet fails due to spatial imprecision or unstable contact dynamics during fine manipulation. This observation inspires us to an alternative paradigm for the assistive robot: rather than re-training or specializing the VLA itself, we preserve the frozen foundation policy and only intervene during localized failure-prone interaction phases. Such a paradigm allows the system to retain the open-world generalization of the underlying VLA while compensating for its lack of contact-level precision through sparse human corrections.

Based on this insight, we propose \emph{Assistron}, a real-time responsive shared autonomy framework that builds upon flow-matching VLA policies without re-training or fine-tuning. Our approach is grounded in two core principles: (1)~minimizing human cognitive and physical effort by leveraging VLA-driven autonomy for macro-movements, and (2)~prioritizing human intervention specifically at critical failure points. Following these principles, Assistron dynamically alternates between autonomous VLA execution and human-in-the-loop intervention using an interaction-aware arbitration mechanism. During intervention, human joystick commands are incorporated as posterior guidance within the flow-matching action generation process. This formulation enables human corrections to remain dynamically consistent with the latent action manifold of the underlying VLA policy, avoiding the instability introduced by naive action-space blending. 
To make posterior-guided shared control responsive despite VLA inference latency, we further develop a real-time execution pipeline \cite{rtc} that combines a Triton-accelerated VLA implementation with real-time chunking adapted to shared control. This allows the next action chunk to be inferred concurrently with execution, preserving both cross-chunk continuity and responsiveness to current user intent.

We validate our framework through a comprehensive, long-horizon \emph{scene recovery} benchmark involving a diverse set of daily tasks, including grasping, placing, inserting, and articulated object manipulation. We compare our system against two baselines: a verbally instructed VLA model and a pure teleoperation setup. Our empirical results demonstrate that neither component alone is sufficient; only by integrating VLA-based semantic priors with continuous human refinement can the system achieve smooth, precise, and effortless assistance, offering a promising direction for scalable assistive manipulation without task-specific retraining.

\begin{figure*}
    \centering
    \includegraphics[width=0.8\linewidth]{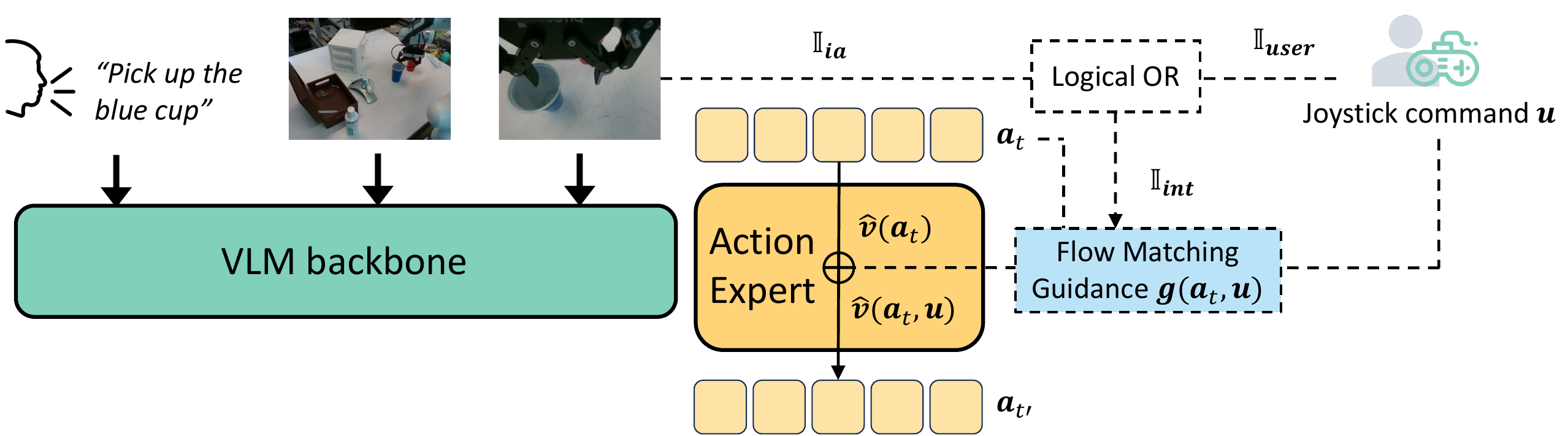}
    \caption{
    Overview of the Assistron. The system employs a VLA model as a general-purpose semantic engine to map verbal language instructions and visual observations into macro-action trajectories. To overcome spatial imprecision during critical interaction phases, the framework introduces an intervention mechanism to solicit the user's intervention. During the intervention, the user's low-level joystick command $\bm{u}$ is continuously fused into the VLA's generative process (Action Expert) as an analytical flow matching guidance term, $\bm{g}(\bm{a}_t, \bm{u})$. 
    }
    \label{fig: assistron}
\end{figure*}

\section{Related Works}
\noindent\textbf{Shared Autonomy.}
Shared autonomy combines human input with autonomous robot assistance to improve the efficiency, safety, and reliability of robot operation. It has been extensively studied in robotic manipulation \cite{javdani2018shared,muelling2017autonomy}, assistive driving \cite{lu2019model}, and wheelchair navigation \cite{psc}. Many prior methods assume a predefined task or a known set of candidate goals, which limits their applicability to open-ended scenarios \cite{javdani2018shared,muelling2017autonomy,yow2023shared}. More recent approaches allow users to express their intent through natural language, enabling robots to provide assistance over a broader range of tasks \cite{liu2025casper,shi2025hirobot,shi2024yay}. Building on this direction, Assistron combines explicit user commands with the action-generation capabilities of a pretrained vision-language-action model.

\noindent\textbf{Intervention Mechanisms and Authority Allocation.}Intervention mechanisms dynamically adjust control authority between a human user and an autonomous policy during task execution. Prior methods commonly estimate task difficulty, policy uncertainty, or failure risk online and shift control toward the human when autonomous execution becomes unreliable. In human-in-the-loop data collection, such mechanisms reduce the need for continuous human supervision by requesting interventions only when necessary \cite{dass2023pato,hoque2022thrifty}. Shared-autonomy systems similarly increase human authority when the autonomous policy is likely to fail \cite{tan2022optimizing,mcmahan2024shared}. These approaches can reduce operator effort while preserving human control at critical moments. In contrast to uncertainty-based intervention, Assistron adjusts authority by identifying interaction moments at which explicit user input should directly influence robot actions.

\noindent\textbf{Vision-Language-Action Models for Shared Autonomy.}Vision-language-action (VLA) models leverage pretrained vision-language representations to condition low-level action generation, transferring broad semantic knowledge to robotic control \cite{intelligence2025pi_}. Recent studies have begun to incorporate VLAs into shared-autonomy systems \cite{cui2025end,liu2026adaptor,ma2026aura,tang2026towards}. However, these approaches typically rely on task-, platform-, or interaction-specific fine-tuning, making them costly to adapt to new robots and user-control modalities. Several recent VLA architectures condition continuous action generators on vision-language embeddings using diffusion or flow matching \cite{ref33,ref35,intelligence2025pi_}. The iterative generation process makes it possible to steer actions at inference time toward additional constraints or objectives \cite{pokle2023training,song2023pseudoinverse,wang2025inference}. Assistron exploits this property through flow-matching guidance, steering the action-generation process toward explicit user commands without fine-tuning the underlying VLA model.

\section{Overview}
\label{sec: overview}

Assistron is a shared autonomy framework that leverages a frozen VLA model to provide seamless assistance for users, particularly those relying on low-bandwidth physical interfaces alongside natural-language commands. As illustrated in Fig.\ \ref{fig: assistron}, Assistron builds upon $\pi_{0.5}$~\cite{intelligence2025pi_} (denoted as $\pi_{\text{vla}}$ in the following text), which utilizes flow matching to map visual observations and linguistic goals directly to action trajectories. We leverage Whisper~\cite{radford2023robust} to transcribe the user's verbal commands into natural-language prompts for the VLA.

To address the local contact-rich failures in zero-shot VLA executions, we formulate a hybrid policy that dynamically toggles between autonomous execution and human-guided refinement. The system policy, $\pi_{\text{sys}}$, is defined as:
\begin{equation}
    \pi_{\text{sys}}(\bm{a}|\bm{s}) = (1 - \mathbb{I}_{\text{int}}) \pi_{\text{vla}}(\bm{a}|\bm{s}) + \mathbb{I}_{\text{int}} \pi_{\text{shared}}(\bm{a}|\bm{s}, \bm{u}), \label{eq: assistron}
\end{equation}
where $\mathbb{I}_{\text{int}} \in \{0,1\}$ is a binary intervention indicator, and $\pi_{\text{shared}}$ denotes the shared control policy that fuses the user's low-level continuous command $\bm{u}$ with the VLA's spatial prior. This formulation strictly reflects our core principles: the first term governs the VLA-driven macro-movements, while the second term activates human intervention specifically at critical failure points. The indicator $\mathbb{I}_{\text{int}}$ determines whether this intervention is required. In the following subsections, we detail the interaction-aware intervention mechanism (Sec.~\ref{sec: intervention}) and the control blending strategy via posterior inference (Sec.~\ref{sec: shared_control}).

\section{Intervention via Interaction Detection} 
\label{sec: intervention}
Pre-trained on an exhaustive amount of data, the VLA has a strong semantic understanding, which helps approach the object to interact with, but it often falters during contact-rich interactions due to spatial imprecision. Minor predictive errors in this phase can lead to critical task failures, such as premature object release or forceful collisions. Therefore, it is crucial to solicit human intervention when the robot transitions from macro-reaching to fine-grained physical interaction. Inspired by the structural consistency of relational manipulation \cite{xue2025demogen, kamil2025sr}, we approximate the interaction phase as the short temporal window preceding a change in gripper contact state (e.g., grasping or releasing). To detect this stage, we introduce a lightweight task-agnostic visual detector $f_{\theta}$, parameterized by a ResNet-18 architecture. Given the wrist-camera observation $I_{\text{wrist}}$, the detector predicts an interaction confidence $p_{\text{it}} = f_{\theta}(I_{\text{wrist}})$. However, visual proximity alone is insufficient for reliable intervention triggering, since the robot may approach objects without initiating contact. To reduce false-positive interventions, we adopt a dual-verification strategy requiring both interaction proximity and an intended gripper actuation predicted by the VLA. The final intervention signal is defined as:
\begin{equation}
    \mathbb{I}_{\text{int}} = \mathbb{I}_{\text{ia}} \lor \mathbb{I}_{\text{user}}, 
    \label{eq: intervention_logic}
\end{equation}
where $\mathbb{I}_{\text{user}}$ denotes manual user intervention, and the autonomous interaction trigger is:
\begin{equation}
    \mathbb{I}_{\text{ia}} = \Big( p_{\text{it}} > \tau_{\text{it}} \Big) \land \Big( |\Delta \tilde{a}_{\text{grip}}| > \delta \Big), 
    \label{eq: auto_trigger} 
\end{equation}
where $\tau_{\text{it}}$ is the interaction probability threshold, and $\Delta \tilde{a}_{\text{grip}}$ denotes the predicted gripper-state change from the VLA policy. When $\mathbb{I}_{\text{ia}} = 1$, the system temporarily suspends the VLA gripper action and transfers control authority to the human operator for fine-grained interaction execution. Once the user completes the grasp or release, control is smoothly returned to the autonomous policy. The system notifies the user of the control-authority transition through on-screen visual feedback and haptic feedback from the joystick. In addition, users may manually intervene at any time through $\mathbb{I}_{\text{user}} = \mathbb{I} (\bm{u} \neq \mathbf{0})$, where $\mathbf{u}$ denotes the human control signal. By unifying phase-aware interaction detection with human intent, this mechanism ensures that the operator only expends effort during high-stakes bottlenecks, thereby preserving task success with a minimum of human effort.

\begin{figure}
    \centering
    \includegraphics[width=0.72\linewidth]{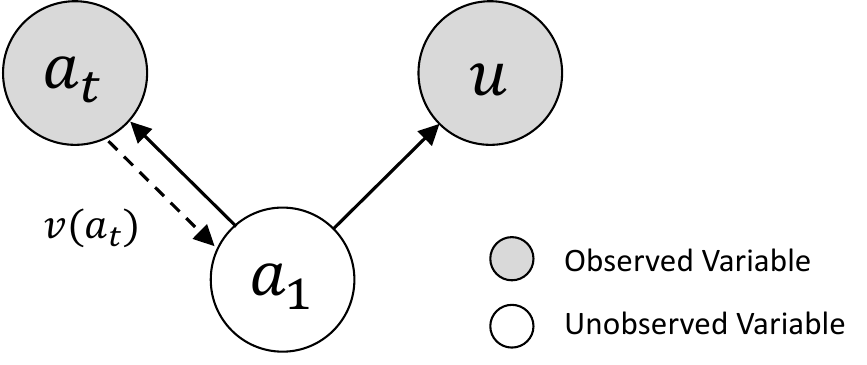}
    \caption{Posterior policy blending. The frozen VLA defines an action prior $p(\bm{a}_1)$, while the user command $\bm{u}$ provides a measurement of the intended action. Flow guidance conditions the generative process toward the posterior $p(\bm{a}_1|\bm{u})$.}
    \label{fig:graphical_model}
\end{figure}

\section{Policy Blending via Posterior Inference}
\label{sec: shared_control}

When intervention is active, i.e., $\mathbb{I}_{\mathrm{int}}=1$, the user provides a low-dimensional joystick command to refine the VLA's behavior during object interaction. Directly executing this command, however, requires the user to control a high-degree-of-freedom robot through a low-bandwidth interface. We therefore treat the human command as a measurement of the intended action and condition the VLA's generative action process on this measurement. This allows the VLA to retain its semantic and motion prior while adapting the generated trajectory to the user's real-time intent.

Our formulation builds on probabilistic shared control \cite{rtv2}, which casts the fusion of human and robot predictions as posterior inference. Let $\bm{a}_1\in\mathbb{R}^{HD}$ denote a vectorized $H$-step action chunk with action dimension $D$. Conditioned on the robot state $\bm{s}$, the frozen VLA defines a prior distribution $\pi(\bm{a}|\bm{s})$. This distribution is represented by a flow-matching model that transforms Gaussian noise $\bm{a}_0\sim\mathcal{N}(\bm{0},\bm{I})$ into an action sample through the probability path $\bm{a}_t = \alpha_t\bm{a}_1+\sigma_t\bm{\epsilon}$, where $\bm{\epsilon}\sim\mathcal{N}(\bm{0},\bm{I})$, $\bm{a}_1$ is the noiseless action, $\alpha_t=t$ and $\sigma_t=1-t$ for $t\in[0,1]$. The instantaneous joystick command is mapped through inverse kinematics to a joint-velocity command and repeated over the action horizon, producing $\bm{u}\in\mathbb{R}^{HD}$. Omitting $\bm{s}$ below for notational simplicity, the desired posterior is
\begin{equation}
    p(\bm{a}_1|\bm{u})
    \propto
    p(\bm{u}|\bm{a}_1)\,p(\bm{a}_1).
    \label{eq:action_posterior}
\end{equation}
We model the measurement likelihood as
$p(\bm{u}|\bm{a}_1)
=\mathcal{N}(\bm{u};\bm{a}_1,\bm{\Sigma}_{\bm{u}})$,
where the diagonal covariance $\bm{\Sigma}_{\bm{u}}$ represents uncertainty in the user's command.

The unconditional vector field $\hat{\bm{v}}(\bm{a}_t,\bm{s})$ that transports Gaussian noise to the VLA action distribution $p(\bm{a}_1)$ can be written as
\begin{equation}
    \hat{\bm{v}}(\bm{a}_t) =  \left( \alpha_{t}\frac{\mathrm{d} \ln(\alpha_{t}/\sigma_{t})}{\mathrm{d}t} \right) \mathbb{E}_{p}[\bm{a}_{1}|\bm{a}_{t}] + \frac{\mathrm{d} \ln \sigma_{t}}{\mathrm{d}t} \bm{a}_{t}.
    \label{eq:unconflow}
\end{equation}
The conditional vector field $\hat{\bm{v}}(\bm{a}_t,\bm{u})$ that generates samples from this posterior $p(\bm{a}_1|\bm{u})$ can be represented as 
\begin{equation}
    \hat{\bm{v}}(\bm{a}_t,\bm{u})  = \left( \alpha_{t}\frac{\mathrm{d} \ln(\alpha_{t}/\sigma_{t})}{\mathrm{d}t} \right) \mathbb{E}_{p}[\bm{a}_{1}|\bm{a}_{t}, \bm{u}] + \frac{\mathrm{d} \ln \sigma_{t}}{\mathrm{d}t} \bm{a}_{t}.
\end{equation}
Considering Tweedie's identity for the Gaussian path:
\begin{equation}
\begin{aligned}
\mathbb{E}[\bm{a}_1|\bm{a}_t,\bm{u}]  =
\mathbb{E}[\bm{a}_1|\bm{a}_t]+\frac{\sigma_t^2}{{\alpha_t}}
\nabla_{\bm{a}_t}\log p(\bm{u}|\bm{a}_t),
\end{aligned}
\label{eq:conditional_tweedie}
\end{equation}
we can substitute Eq.~\eqref{eq:conditional_tweedie} into the conditional flow, which yields:
\begin{align}
& \hat{\bm{v}}(\bm{a}_t,\bm{u}) =\hat{\bm{v}}(\bm{a}_t)+\bm{g}(\bm{a}_t,\bm{u}) \\
& \bm{g}(\bm{a}_t,\bm{u})=
    \sigma_t^2
    \frac{\mathrm{d}\log(\alpha_t/\sigma_t)}{\mathrm{d}t}
    \nabla_{\bm{a}_t}\log p(\bm{u}|\bm{a}_t).
\label{eq:general_flow_guidance}
\end{align}

To evaluate the likelihood score efficiently, we adopt the local Gaussian approximation $p(\bm{a}_1|\bm{a}_t)
    \approx
    \mathcal{N}
    \left(
        \bm{a}_1;
        \hat{\bm{a}}_1,
        r_t^2\bm{I}
    \right),
p(\bm{u}|\bm{a}_t)
    \approx
    \mathcal{N}
    \left(
        \bm{u};
        \hat{\bm{a}}_1,
        r_t^2\bm{I}+\bm{\Sigma}_{\bm{u}}
    \right),
$
where $\hat{\bm{a}}_1$ is a one-step estimate of the clean action chunk as
$
    \hat{\bm{a}}_1
    =
    \bm{a}_t+(1-t)\hat{\bm{v}}(\bm{a}_t)$, and $r_t^2=\frac{\sigma_t^2}{\sigma_t^2+\alpha_t^2}$.
Defining
$\bm{C}_t=r_t^2\bm{I}+\bm{\Sigma}_{\bm{u}}$ and
$\bm{J}_t=\partial\hat{\bm{a}}_1/\partial\bm{a}_t$, the likelihood score is
\begin{equation}
\nabla_{\bm{a}_t}\log p(\bm{u}|\bm{a}_t)
    \approx
    \bm{J}_t^{\mathsf{T}}
    \bm{C}_t^{-1}
    (\bm{u}-\hat{\bm{a}}_1).
    \label{eq:user_likelihood_score}
\end{equation}
Computing $\bm{J}_t$ requires differentiating through the VLA action expert at every flow step. To enable real-time control, we use the identity-Jacobian approximation $\bm{J}_t\approx\bm{I}$. 
Substituting this approximation together with $\alpha_t=t$ and $\sigma_t=1-t$ into Eq.~\eqref{eq:general_flow_guidance} gives the analytical guidance:
\begin{equation}
    \bm{g}(\bm{a}_t,\bm{u})
    \approx
    \frac{1-t}{t}
    \left(
        \frac{(1-t)^2}{(1-t)^2+t^2}\bm{I}
        +\bm{\Sigma}_{\bm{u}}
    \right)^{-1}
    (\bm{u}-\hat{\bm{a}}_1).
\label{eq:guidance3}
\end{equation}
The covariance $\bm{\Sigma}_{\bm{u}}$ controls the relative authority of the human command: smaller diagonal entries enforce closer agreement with $\bm{u}$, whereas larger values preserve more of the VLA prior. By integrating the guided vector field $\hat{\bm{v}}(\bm{a}_t,\bm{u})$, Assistron approximately samples from $p(\bm{a}_1|\bm{u})$ while keeping the underlying VLA frozen. The user command therefore guides the entire action-generation process instead of being blended with the VLA prediction only after generation.

\begin{figure}
  \centering
  \includegraphics[width=\linewidth]{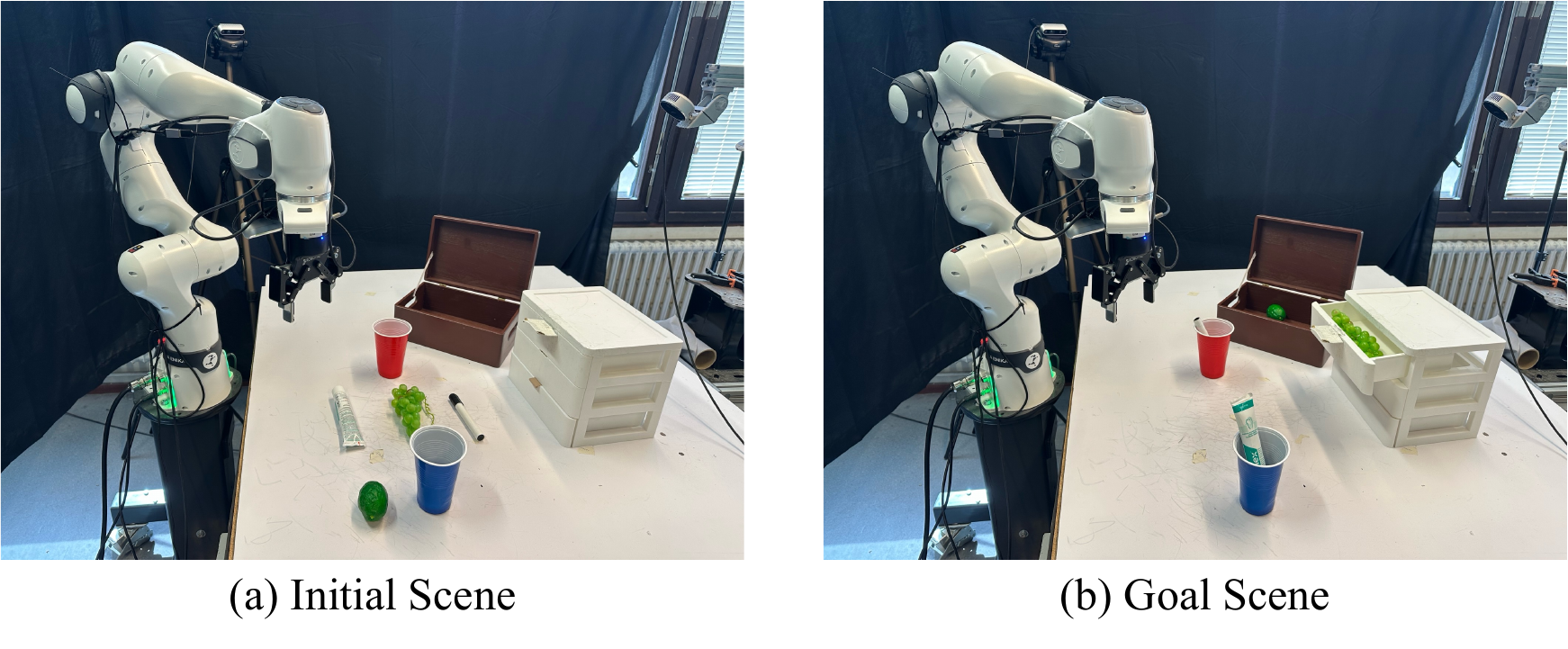}
  \caption{The task setup for the scene recovery experiment. At the beginning of the experiment, the table will be set as the Initial Scene (a). The users are shown the Goal Scene (b), and asked to recover the Goal Scene by controlling the robot.}
  \label{fig: exp setup}
  \vspace{-0.2cm}
\end{figure}


\noindent\textbf{Real-Time Shared-Control Execution.} To realize a smooth shared control, we implement a Triton-implemented $\pi_{0.5}$ \cite{ma2025running}, which can achieve approximate 60\,ms latency compared to 150\,ms latency on the original JAX implementation on Nvidia RTX A6000. Though accelerated, synchronous replanning remains limited to 8\,Hz and can introduce blocking gaps between consecutive action chunks. We therefore adapt real-time chunking \cite{rtc} to posterior-guided shared control. At replanning step $k$, let $\hat{\bm a}^{k-1}$ denote the previous action chunk and $d_k$ the number of actions expected to be consumed while the next inference request is processed, estimated conservatively from recent inference delays. We construct a reference chunk by concatenating the unexecuted prefix of $\hat{\bm a}^{k-1}$ with the latest user command $\bm{u}^k$:
\[
\tilde{\bm u}_i^k =
\begin{cases}
\hat{\bm a}^{k-1}_{s_k+i}, & i<d_k,\\
\bm u^k_{i-d_k}, & i\ge d_k,
\end{cases}
\]
where $s_k$ is the number of actions already consumed from the previous chunk. The overlapping prefix preserves temporal continuity, while the remaining horizon encodes the user's current intent. We then condition the flow-matching process on $\tilde{\bm u}_i^k$ through posterior guidance. Because inference runs concurrently with execution, the first $d_k$ actions of the newly generated chunk correspond to the elapsed inference interval and are discarded; only the remaining suffix is submitted for execution. This design allows the robot to continue moving while generating the next posterior-guided chunk, increasing the effective shared-control update rate from 8\,Hz with synchronous Triton inference to 15\,Hz with Triton and real-time chunking, an approximate $2\times$ improvement.

\begin{figure*}[!tb]
    \centering
    \includegraphics[width=\linewidth]{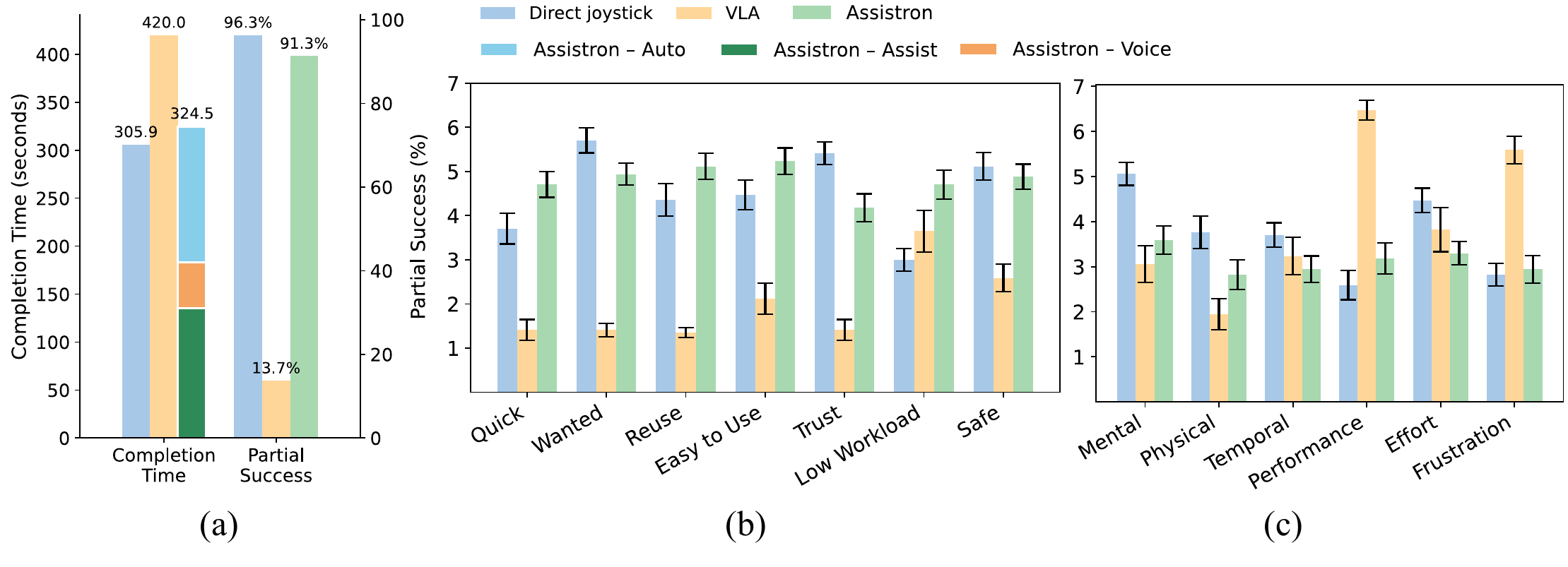}
    \caption{Experiment results of the scene recovery task. (a) Objective performance. (b) The user satisfaction survey. (b) The NASA-TLX survey. For (b) a higher score is better, while for (c) a lower score is better. 
    }
    \label{fig: results}
\end{figure*}

\begin{figure*}[!tb]
    \centering
    \includegraphics[width=0.8\linewidth]{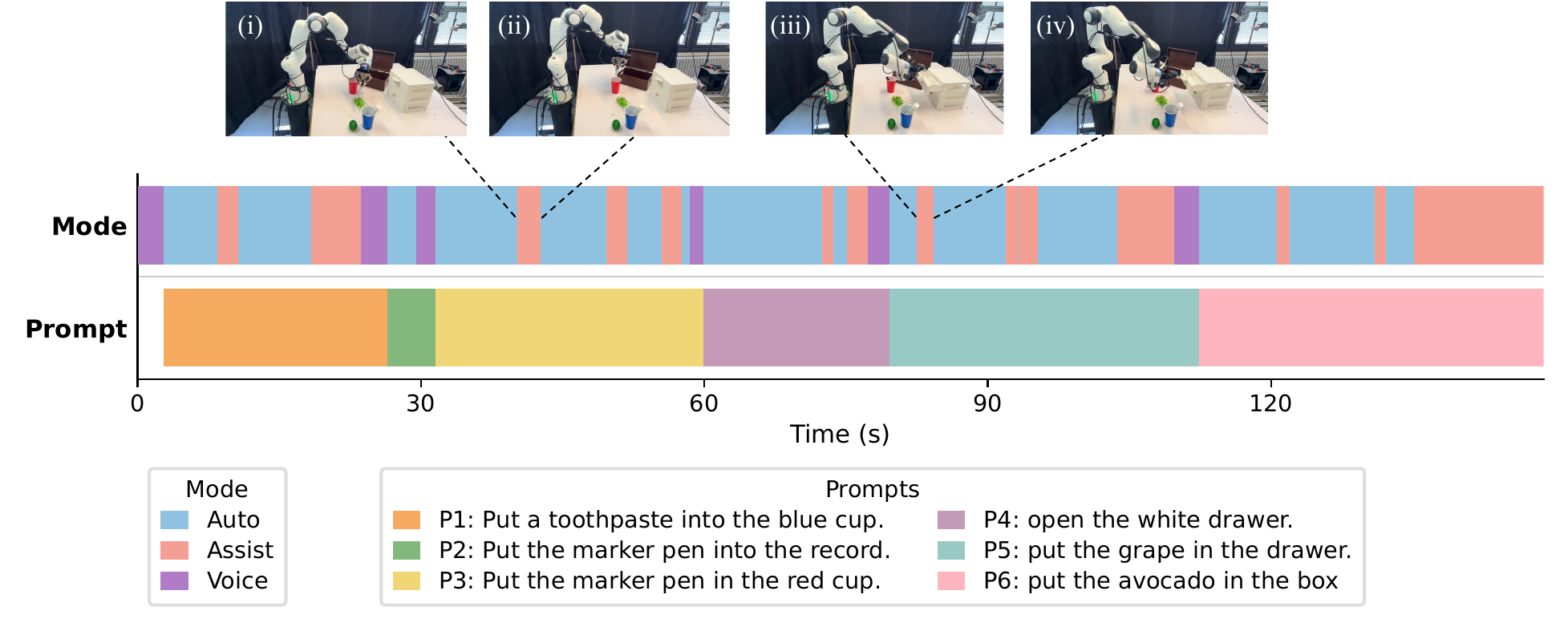}
    \caption{
     Visualization of the scene recovery task. The top sequence visualizes critical keyframes (i)--(iv) corresponding to moments of human intervention. The middle timeline illustrates the dynamic transitions of the control state (\emph{Auto}, \emph{Assist}, or \emph{Voice}) over the duration of the execution. The bottom timeline indicates the active natural language prompt (P1--P6) guiding the VLA at any given time. See details in the text.
    }
    \label{fig: demo_full}
\end{figure*}

\section{Experiments}
In this section, we evaluate the applicability of Assistron through three research questions. \textbf{RQ1} (Sec.~\ref{sec: scene recovery}) addresses Assistron's overall performance, asking \textsl{``Does \emph{Assistron} improve task performance and user experience compared to existing methods?''} \textbf{RQ2} (Sec.~\ref{sec: policy blending}) studies Assistron's flow-matching guidance mechanism, and asks \textsl{``Does the posterior policy blending outperform other alternatives during the intervention?''}  \textbf{RQ3} (Sec.~\ref{sec: interaction detection}) evaluates Assistron's interaction detection model and quantifies its ability to effectively detect interaction phases.


\subsection{Real-World Shared Autonomy Performance and User Study} \label{sec: scene recovery}

Here, we comprehensively study our method's applicability and performance in a long-horizon task.

\noindent \textbf{Design}. We consider a \emph{scene recovery} task where the user is expected to rearrange objects to match a specified configuration. The goal configuration is specified via an image. Starting from an initial configuration, the user issues verbal and joystick commands to the robot to move objects to their goal pose. Because the robot is unaware of the goal configuration, it must continuously infer the user's intended actions from its multimodal inputs. To keep experiments within a reasonable time budget, any run exceeding seven minutes is automatically terminated and marked as failed. We compare the proposed method with two baselines. The first baseline uses neither assistance nor VLA, giving the user full authority via joystick commands (referred to as ``Direct joystick''). The second baseline consists of a VLA that has full authority to execute motor commands derived from verbal user commands (referred to as ``VLA'').


\noindent \textbf{Task Description}.  As illustrated in Fig.~\ref{fig: exp setup}, the scene recovery task comprises five sub-tasks: (a) opening the drawer; (b) placing the grape into a drawer; (c) placing the avocado into a box; (d) inserting the marker pen into the red cup; and (e) inserting the toothpaste into the blue cup. These sub-tasks encompass fundamental daily manipulation skills—such as pick-and-place, insertion, and pulling—serving as representative proxies for broader everyday activities. Crucially, users are not provided with explicit verbal scripts for these tasks, requiring them to formulate their own instructions. 


\noindent \textbf{Protocol}. We enrolled 17 novice participants from the local community (14 male, 3 female; age range: 18–40). Participants completed a practice session before completing tasks with each method in randomized order. After each, they answered the user satisfaction and NASA-TLX survey \cite{nasatlx}. All participants provided signed consent for the experimental procedure, which was approved by the Social and Societal Ethics Committee of (anonymous institution). The collected data were processed in accordance with the General Data Protection Regulation (GDPR) of (anonymous organization).


\noindent \textbf{Objective Performance.} Figure~\ref{fig: results}~(a) compares the methods across \emph{completion time} and \emph{partial success} (20\% per sub-task). VLA consistently timed out. \emph{Assistron} (324.5s) took slightly longer than \emph{Direct Joystick} (305.9s) due to the VLA's conservative macro-reaching speeds. However, \emph{Assistron} required active user input for only 56.5\% of the duration (41.7\% joystick, 14.8\% voice), while autonomously handling the remaining 43.5\%. This halves the active control time compared to \emph{Direct Joystick}, substantially reducing continuous human effort. For \emph{partial success}, spatial imprecision limited VLA to 13.7\%. \emph{Assistron} achieved 91.3\%, closely trailing \emph{Direct Joystick} (96.3\%). This slight gap occurred because occasional VLA semantic confusion required users to manually override misaligned actions. We also observe that the user who has poorer \emph{Direct Joystick} performance will have more improvement in completion time when assisted by \emph{Assistron} (Pearson correlation $r=0.762, p=0.001$), indicating that less proficient users derive substantially greater benefits from the system.

\noindent \textbf{Subjective Evaluation.} Figure~\ref{fig: results}~(b-c) presents the user satisfaction and NASA-TLX results. We analyzed main effects across control methods (\emph{Direct Joystick}, \emph{VLA}, \emph{Assistron}) using the Friedman test, followed by Wilcoxon signed-rank tests with Benjamini-Hochberg correction for post-hoc comparisons. In the satisfaction survey, all metrics showed significant effects ($p < 0.05$). \emph{VLA} scored lowest on positive metrics due to poor task completion. Compared to \emph{Direct Joystick}, \emph{Assistron} scored significantly higher in being \emph{Quick}, \emph{Easy to Use}, having \emph{Low Workload}, and \emph{Reuse} ($p < 0.05$), but lower in \emph{Wanted} and \emph{Trust} ($p < 0.05$). This indicates \emph{Assistron} improves efficiency and ease at the cost of some absolute trust and direct agency.  For the NASA-TLX assessment, significant effects emerged in \emph{Mental}, \emph{Physical}, \emph{Performance}, and \emph{Frustration} ($p < 0.05$). While \emph{VLA} required the least \emph{Physical} effort, it yielded the highest \emph{Frustration} and poorest \emph{Performance}. Crucially, \emph{Assistron} demanded significantly less \emph{Mental} and \emph{Physical} effort than \emph{Direct Joystick} ($p < 0.001$). This workload reduction corroborates the satisfaction ratings, confirming \emph{Assistron}'s effectiveness in providing effortless shared autonomy. Furthermore, correlation analysis between \emph{Teleop}'s completion time and subjective score differences ($\Delta = \text{Assistron} - \text{Teleop}$) reveals that less proficient users experienced a significantly greater reduction in frustration ($r = -0.564, p = 0.023$).

\noindent \textbf{Visualization of Shared Autonomy.} Fig.~\ref{fig: demo_full} illustrates a demonstration of how the human user collaborates with the VLA in our method. We highlight the distinct operating modes—autonomous VLA execution $\pi_{\text{vla}}(\bm{a}|\bm{s})$ (\emph{Auto}), the shared policy $\pi_{\text{shared}}(\bm{a}|\bm{s}, \bm{u})$ (\emph{Assist}), and verbal command recording (\emph{Voice})—along with the transitions between them. The corresponding language prompts used throughout the process are also displayed. During the entire task, the \emph{Auto} mode accounts for over 50\% of the execution time, demonstrating the VLA's ability to significantly reduce user workload. Conversely, user intervention remains crucial for successful task completion. For instance, in keyframe (i), the VLA failed to locate the red cup via its wrist camera after picking up the pen, and erroneously attempted to drop it. At this juncture, the system transitioned to \emph{Assist} mode, allowing the user to intervene and manually guide the pen directly above the red cup (keyframe (ii)). In another instance, after opening the drawer and receiving the new verbal command, ``put the grape in the drawer,'' the VLA remained fixated on the drawer instead of retracting to grasp the grape. The user promptly assisted the VLA in retreating from the drawer, enabling the robot to approach the grape afterward. This demonstration underscores that seamlessly combining the VLA's autonomous capabilities with targeted human intervention achieves optimal performance while minimizing continuous human effort.



\begin{figure}
  \centering
  \includegraphics[width=\linewidth]{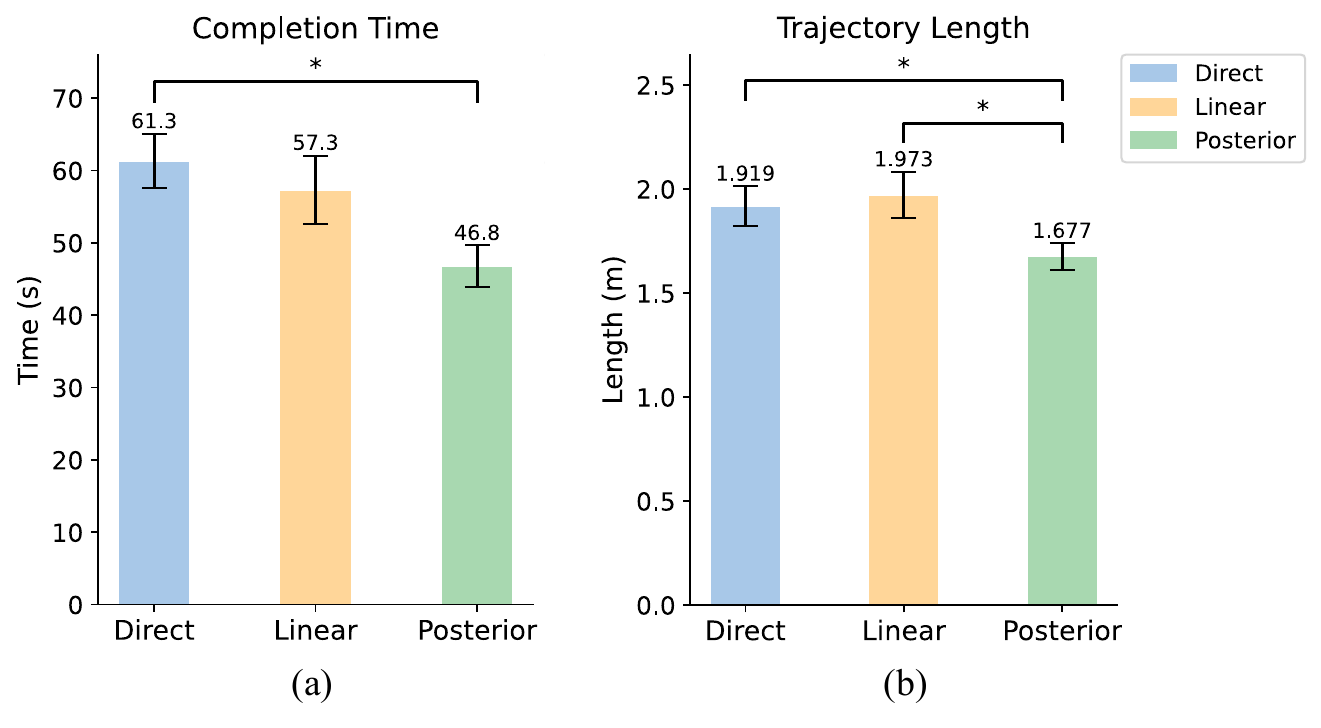}
  \caption{Comparison of completion time (a) and trajectory length (b) across different control methods. \emph{Posterior} blending significantly improves performance over the baselines.}
  \label{fig: ablation}
\end{figure}

\begin{figure*}[!tp]
    \centering
    \includegraphics[width=\linewidth]{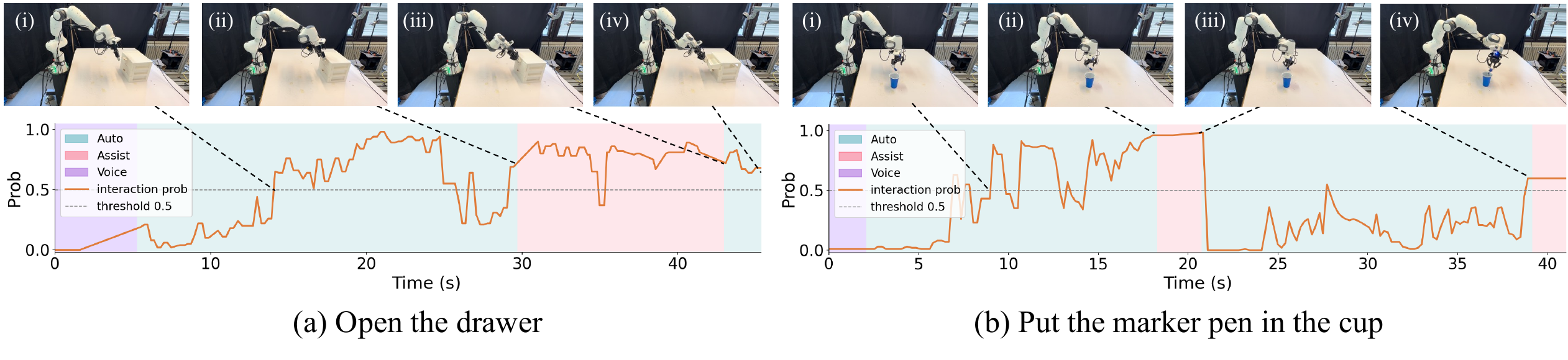}
    \caption{
    Temporal visualization of the interaction detection and the transitions of the control state (\emph{Auto}, \emph{Assist}, or \emph{Voice}) over the duration of the execution
    The figure illustrates the predicted interaction probability $p_{\text{it}}$ and the demonstrations of two representative scene recovery tasks: (a) opening a drawer and (b) putting a marker pen into a cup. Keyframes (i)-(iv) correspond to critical moments.
    }
    \label{fig: demos}
    \vspace{-0.2cm}
\end{figure*}

\subsection{Ablation Study on Policy Blending} \label{sec: policy blending}

To evaluate the proposed posterior policy blending framework, we conduct an ablation study on a single task—placing a \emph{grape into a drawer}. The VLA is conditioned on the prompt ``Open the drawer, and put the grape in the drawer,'' and only assists the robot when the user provides a low-level command. We compare our proposed \emph{Posterior} blending with linear blending (\emph{Linear}) and pure teleoperation (\emph{Direct}). Each participant completed the task twice per method in a randomized, anonymous order after a practice session.

As shown in Fig.~\ref{fig: ablation}, \emph{Posterior} significantly reduces completion time compared with \emph{Direct} and produces significantly shorter trajectories than both \emph{Direct} and \emph{Linear} ($p<0.05$), based on the Friedman test followed by pairwise Mann--Whitney U tests with a two-stage Benjamini--Hochberg correction. Blending generally improves efficiency by assisting multiple robot degrees of freedom simultaneously. However, linear blending can become ineffective when the flow-matching VLA produces multimodal actions that conflict with the user's command. By explicitly accounting for the VLA action distribution, \emph{Posterior} provides more consistent and user-aligned assistance.


\noindent\textbf{Comparison with Diffusion Assistance.}
Diffusion assistance incorporates user input by initializing the denoising process at an intermediate timestep with a noised command,$\bm{a}{t'}=(1-\gamma)\bm{u}+\gamma\bm{\epsilon}$, where $t'=1-\gamma$ and $\bm{\epsilon}\sim\mathcal{N}(\bm{0},\bm{I})$ \cite{yoneda2023noise,luo2025human}. This construction assumes that the user command specifies the same action space as the generative policy. This assumption does not hold for $\pi{0.5}$ fine-tuned on DROID. At each horizon step, the flow model jointly denoises a 32-dimensional vector $[\bm{a}^{\mathrm{exec}}_t,\bm{d}_t]$: only the 8-dimensional $\bm{a}^{\mathrm{exec}}_1$ is executed by the robot, whereas the remaining 24 dimensions $\bm{d}_1$ are discarded only after generation. Although not executed, $\bm{d}_t$ participates in every flow-matching step and is jointly modeled with $\bm{a}^{\mathrm{exec}}_t$. The user command can therefore initialize $\bm{a}^{\mathrm{exec}}_{t'}$ but provides no corresponding value for $\bm{d}_{t'}$. Independently filling these unspecified dimensions with Gaussian noise produces a hybrid initialization that disrupts the learned cross-dimensional structure, resulting in erratic action chunks. In contrast, our method starts from a complete sample from the model prior and incorporates the user command as posterior guidance throughout the flow, allowing all 32 dimensions to evolve coherently. On the same drawer task, posterior guidance reduced the median acceleration RMS from 12.30 to 2.97 rad/s$^2$ ($4.1\times$), jerk RMS from 301.23 to 44.94 rad/s$^3$ ($6.7\times$), and inter-chunk discontinuity from 1.039 to 0.357 rad/s ($2.9\times$). These results indicate that, for this partially executed action representation, posterior guidance provides substantially smoother assistance than directly adapting diffusion assistance through a partially specified initialization.

\subsection{Interaction Detection Performance} \label{sec: interaction detection}
We evaluate the proposed task-agnostic interaction detector on a dataset containing over 12k wrist-camera frames collected from diverse tebletop tasks, including grasping, placing, insertion, and articulated manipulation. Ground-truth interaction labels are automatically extracted, where frames within a 2-second temporal window preceding a gripper state change are labeled as positive interaction samples. Trained solely on $224 \times 224$ wrist observations, our model achieves 81.2\% test accuracy and 84.5\% average precision (AP). The high AP underscores the model's robust recall of positive instances, ensuring the reliable anticipation of critical task bottlenecks. 

\textbf{Visualization of Phase Transitions.} Fig.~\ref{fig: demos} illustrates the dynamic control shifts driven by the predicted interaction probability ($p_{\text{it}}$). In the \textbf{drawer-opening task} (Fig.~\ref{fig: demos}~(a)), the VLA smoothly executes the macro-reaching motion (\emph{Auto} phase). At keyframe (i), visual proximity to the handle drives $p_{\text{it}}$ above the 0.5 threshold. However, the system transitions to shared control (\emph{Assist} phase) at keyframe (ii) only when the VLA actively initiates a grasp ($|\Delta \tilde{a}_{\text{gripp}}| > \epsilon$). Because the VLA's predicted pose is misaligned, the user briefly intervenes to correct the end-effector and secure the handle (ii-iii). Once the user completes the grasp and human actuation stops, the VLA instantly regains authority at (iii) to autonomously pull the drawer (iv). The \textbf{pen insertion task} (Fig.~\ref{fig: demos}~(b)) demonstrates a multi-phase handover. Following a similar initial user intervention to properly grasp the pen (ii), the VLA resumes control for the macro-transport, navigating the object to a vertical position above the cup (iii). Crucially, a second interaction phase is triggered at (iv) just as the VLA prepares to open its gripper. Control shifts back to the user, enabling a precise, manual release that successfully concludes the task without the risk of a premature drop.



\section{Conclusion}
In this work, we present Assistron, a shared-autonomy framework that leverages a frozen VLA to assist users in activities of daily living. Rather than re-training or specializing the VLA, Assistron exploits the VLA's semantic understanding for macro-reaching and solicits the user's intervention for the localized contact-rich interaction, which preserves the generalization capability of the VLA and compensates for local precision failures. During intervention, users' commands are incorporated as posterior guidance within the flow-matching action generation process, enabling human corrections to remain dynamically consistent with the latent action manifold of the underlying VLA policy. We validated our approach through a comprehensive scene recovery benchmark encompassing a diverse range of daily manipulation tasks. Empirical results demonstrate that by fusing semantic priors with timely human refinement, Assistron significantly improves task success rates while minimizing operator workload—offering a promising direction for scalable assistive manipulation without task-specific fine-tuning or re-training. Building on this foundation, future work will evaluate its accessibility and real-world benefits through co-design and studies with people with motor impairments and wheelchair users.

\noindent \textbf{Limitations}
\emph{(1) Dependence on VLA capabilities.} 
Assistron relies on a frozen VLA policy to provide semantically meaningful macro-actions. As a result, the overall system remains fundamentally bounded by the capabilities of the underlying VLA. Severe semantic failures, such as incorrect instruction grounding or navigation toward irrelevant regions, cannot be reliably corrected through local shared-control intervention. 
\emph{(2) Assumption on user-policy alignment.} 
Our shared-control formulation (\ref{sec: shared_control}) implicitly assumes that user corrections lie within the action distribution induced by the VLA policy. In practice, human actions may deviate from this distribution, potentially causing conflicts between user intent and VLA priors. Future work may incorporate out-of-distribution detection to identify such cases and fall back to pure teleoperation when necessary. 
\emph{(3) Limited evaluation scope.}
Although our ultimate goal is to assist people with motor impairments, the current study included only able-bodied novice participants. Consequently, the findings may not fully generalize to the target population. This study serves as a preliminary evaluation, and future work will validate and adapt the framework through studies involving participants with motor impairments.

\bibliographystyle{IEEEtran}
\bibliography{ref}

\end{document}